\documentclass{article}
\usepackage[eatrack]{log_2024}		

\usepackage{booktabs}						
\usepackage{multirow}						
\usepackage{amsfonts}						
\usepackage{graphicx}						
\usepackage{duckuments}						

\usepackage[numbers,compress,sort]{natbib}	
\usepackage{svg}
\usepackage{pifont}

\title[Enriching GNNs with Contextual Representations for Detecting Disinformation on Social Media]{Enriching GNNs with Text Contextual Representations for Detecting Disinformation Campaigns on Social Media}

\author[B. Silva et al.]{%
Bruno Croso Cunha da Silva\thanks{Equal contribution.}
\\
Universidade de São Paulo, Brazil\\
\email{croso.bruno @usp.br}\And
Thomas Palmeira Ferraz\footnotemark[1]\\
Institut Polytechnique de Paris, France\\
\email{thomas.ferraz @alumni.usp.br}\And
Roseli de Deus Lopes\\
Universidade de São Paulo, Brazil\\
\email{roseli.lopes @usp.br}
}

\begin{document}

\maketitle

\begin{abstract} 
Disinformation on social media poses both societal and technical challenges, requiring robust detection systems. While previous studies have integrated textual information into propagation networks, they have yet to fully leverage the advancements in Transformer-based language models for high-quality contextual text representations. This work addresses this gap by incorporating Transformer-based textual features into Graph Neural Networks (GNNs) for fake news detection. We demonstrate that contextual text representations enhance GNN performance, achieving 33.8\% relative improvement in Macro F1 over models without textual features and 9.3\% over static text representations. We further investigate the impact of different feature sources and the effects of noisy data augmentation. We expect our methodology to open avenues for further research, and we made code publicly available.\footnote{Code available at: \url{https://github.com/BrunoCroso/ContextualGNNs-FakeNews}}
\end{abstract}

\section{Introduction}

The spread of fake news on social media poses a serious societal challenge, disrupting public opinion and undermining trust in the media. While progress has been made in fake news detection using language processing \cite{dEFEND,nasir2021fake} and hierarchical graph propagation \cite{shu2020hierarchical} separately, recent works have yet to fully exploit their combined potential to develop graph-based models that capture both the structural and semantic properties of social media networks. Graph Neural Networks (GNNs) \cite{GNN_original2008} are particularly well-suited for this task, given their ability to model complex information propagation. However, the noisy, incomplete nature of social media data, including user interactions and profile details, still poses significant challenges.

Despite the suitability of GNNs for this domain, integrating advanced textual features into propagation networks remains underexplored. A few studies have incorporated textual information into hierarchical graph propagation tasks \cite{gcan,hamid2020fake}, including static text representations with GNNs for fake news detection \cite{michail2022detection}. However, these approaches overlook the capabilities of recent Transformer-based language models (LMs), which provide high-quality contextual representations by capturing complex semantic relationships \cite{bert2019,nguyen2020bertweet,reimers-gurevych-2019-sentence}. Furthermore, there is limited systematic evaluation of how textual features affect node representations within GNNs, resulting in a gap in understanding their role in disinformation detection.

To advance this approach, we investigate how incorporating textual information from user profiles (bios) and user interactions (retweets) affects GNN performance in detecting disinformation campaigns on Twitter (X). We hypothesize that certain behavioral patterns, such as those from bots or biased profiles, can be better captured by incorporating text into the propagation graph. Our contributions include systematically evaluating both static and contextual text representations in the graph setting, and addressing the class imbalance challenge—a common obstacle in fake news detection. Specifically, we explore Noisy Embedding Augmentation \cite{jain2023neftune}, a technique widely applied in the text domain to enhance robustness through simulated data perturbations, and assess its effectiveness in GNN training.

Our results demonstrate that incorporating contextual text representations into GNNs improves performance significantly, achieving relative Macro F1 gains of 9.3\% over static text representations and 33.8\% over GNNs without text. Retweet content provided richer contextual signals than profile content, but combining both yielded the best overall performance. Conversely, noise injection on textual features caused instability during training, reducing performance and making it unsuitable for GNN tasks. To the best of our knowledge, this is the first work to systematically explore the integration of contextual text representations into GNNs for fake news detection. By combining contextual textual features from LMs with graph-based modeling, our study advances the understanding of their role in disinformation detection. We also contribute a methodology for evaluating architectural changes in GNNs, encouraging further research in this domain. Code is made publicly available\footnotemark[2].
\begin{figure}
    \centering
    \includegraphics[width=\textwidth]{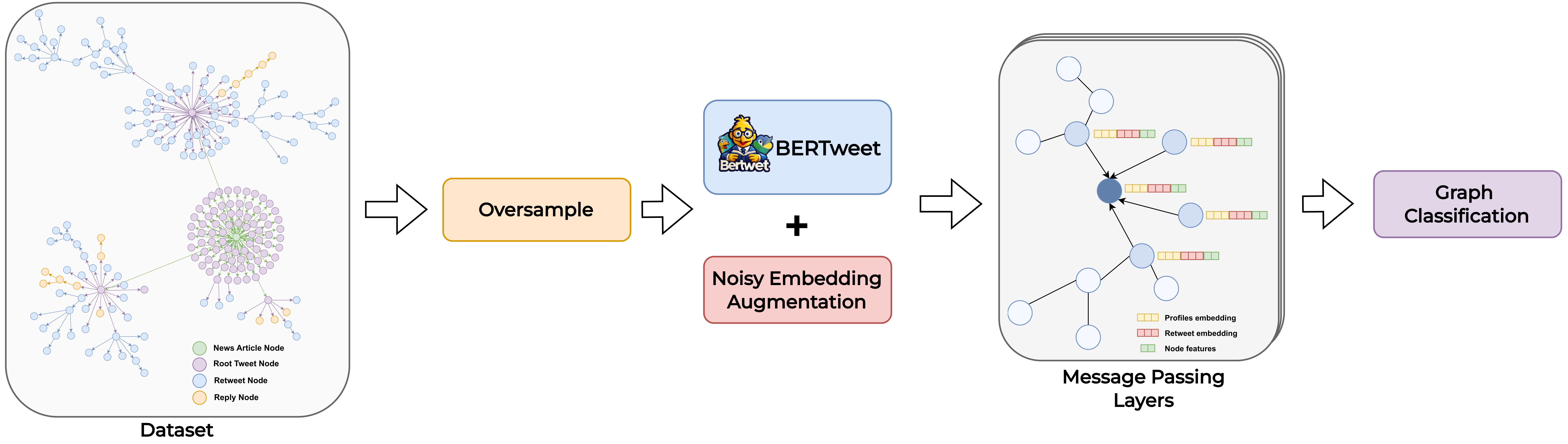}
    \caption{The pipeline of our Text-Enriched GNNs starts with propagation graphs where the initial node represents a news article, and subsequent nodes form merged diffusion trees of root tweets, retweets, and replies. The dataset is oversampled to address class imbalance. Node features are enriched with textual embeddings from user profiles and retweets using BERTweet, with optional noise augmentation via NEFTune. Message-passing layers with pooling aggregate the nodes into graph-level a representation for producing a classification about the news article.}
    \label{fig:diagram}
    \vspace{-8pt}
\end{figure}
\section{Proposed Model}
\paragraph{GNNs for Disinformation Campaign Detection.} Detecting disinformation campaigns involves classifying the propagation network $\mathcal{G}$ of a news article item $v_0$ shared on social media. This network is built by merging all diffusion trees $\vec{Dt_{i}} = \{v_{i0}, \{v_{i10}, \{ v_{i11}, \dots\}, \dots\}, \dots\} $, obtained from all $v_{i0}$ root publications mentioning the news $v_0$, following \citet{shu2020hierarchical} and \citet{michail2022detection}. The resulting directed graph $\mathcal{G} = (V, E)$ represents the radial spread of information, with $v_0$ as the central node and edges tracing the social media diffusion process. The task is framed as a binary classification problem to determine whether $\mathcal{G}$ corresponds to fake or true news.

To achieve this, we construct a feature vector $\vec{X_v}$ for each node $v \in G$, encoding its key characteristics. Graph Neural Networks (GNNs) leverage message-passing to update $\vec{X_v}$ with contextual information from neighboring nodes. We use Graph Attention Networks (GATs) \cite{velickovic2018graph}, which employ attention mechanisms for neighbor aggregation, producing node embeddings. These embeddings are pooled to form a graph-level representation, and passed to a Multi-Layer Perceptron (MLP) for classification. Figure \ref{fig:diagram} illustrate this process. 
\vspace{-8pt}

\paragraph{Text Representations.} To enrich the feature vectors $\vec{X_v}$ of nodes $v \in G$, we incorporate either static or contextual textual embeddings. Static encoders assign fixed embeddings to words, regardless of context. For example, the word "bank" will always have the same embedding, whether referring to a financial institution or a riverbank. Contextual encoders, on the other hand, generate dynamic embeddings based on surrounding words, capturing nuanced meanings in specific contexts.
\vspace{-4pt}
\paragraph{Imbalance Problem} 
A typical issue in fake news detection is the highly negative class imbalance problem, where there is significantly more true news than fake news in the real world. Depending on the goal, the detector may prioritize flagging relevant instances for human review (recall) or ensuring only truly fake news is flagged (precision). Balancing these objectives is critical and raises ethical concerns \cite{ferraz2024explainable}.

Classical methods like downsampling (reducing majority class data) or oversampling (duplicating minority instances) are often inadequate for graph classification. Downsampling risks losing valuable structural data, while oversampling risks overfit to specific graph structures, as each graph instance should typically be unique. However, oversampling still has shown promise in graph classification tasks \cite{michail2022detection}.

Recent methods address imbalance through graph-tailored data augmentation, such as structural manipulation or feature noise injection \cite{liu2022local,gao2019graph,zhou2023graphsr,zhao2024imbalanced,you2020graph}. Adding noise to node features has improved robustness across tasks \cite{liu2022local,kong2021flag}, and similar trends are observed in text processing, where noisy embedding augmentation enhanced classification \cite{jain2023neftune,liang2024drbert,chen2024slight,classificationneftune2024}. Building on this, we investigate combining oversampling with noise injection into textual features.
\vspace{-4pt}
\paragraph{Research Questions.} In this work, we examine the impact of incorporating textual content into the social media propagation network on the performance of GNNs for disinformation campaign detection. Specifically, considering the class imbalance problem, we investigate the following research questions: \textit{\textbf{RQ1.} Does incorporating textual content from the user profiles involved in spreading the news (referred to as \underline{profiles}) improve the model’s performance? \textbf{RQ2.} Does incorporating textual content from user interactions in the propagation process (referred to as \underline{retweets}) enhance the model’s performance? \textbf{RQ3.} How does the choice between \underline{static} and \underline{contextual} text representations for these textual features affect performance? \textbf{RQ4.} Can oversampling with \underline{feature noise injection} improve model convergence and overall performance?}

\section{Empirical Setup}


\paragraph{Dataset.} We conduct experiments on the Politifact subset of the FakeNewsNet dataset \cite{shu2020fakenewsnet}, which focuses on U.S. political news. This dataset provides news propagation networks and user comment histories, labeled by the homonymous fact-checking service. We use diffusion trees from \citet{michail2022detection} and enhance it with additional retweet and profile data collected via the Twitter API\footnote{Data collection done before X's new policies restricted API access.}, removing any samples deleted from the platform. The final dataset includes 1,242 fake news and 10,793 true news propagation graphs, reflecting an 11.5\% imbalance ratio.
\vspace{-4pt}
\paragraph{Graph Model.} Our model consists of  two GAT layers: the first with 32 units and 4 attention heads, and the second with a single head, followed by a graph pooling layer that averages node embeddings. This setup is inspired by \citet{michail2022detection}, but we remove GraphSAGE, retaining only the message-passing layer as the aggregation mechanism, to avoid potential confounding interactions with noisy data augmentation and different text representations\footnote{For instance, the different embedding sizes could impact GraphSAGE performance, making configurations not comparable. Reintroducing GraphSAGE could be explored in future work.}. Each node’s feature vector is defined as $\vec{X_v} = [\vec{x_{v_{1}}}, \vec{x_{v_{2}}},  \vec{x_{v_{3}}}]$, where $[,]$ represents concatenation.  Here, $\vec{x_{v_{1}}}$ contains propagation-related features (e.g. user attributes like follower count, delay between the post and its predecessor, etc.), while $\vec{x_{v_{2}}}$ and $\vec{x_{v_{3}}}$ represent the textual features from user profiles and retweets, respectively, if used.

\vspace{-4pt}
\paragraph{Text Embeddings.} We compare the static encoder GloVe \cite{pennington2014glove} and the Transformer-based contextual encoder BERTweet \cite{nguyen2020bertweet}, pre-trained only on Twitter data. Textual features $\vec{x_{v_2}}$ and $\vec{x_{v_3}}$ have dimensions of 100 (GloVe) or 768 (BERTweet), computed as the average of the word embeddings. For both models, we assess the impact of including textual content from user profiles (\textit{profiles}) and user interactions (\textit{retweets}). We also explore augmenting text features with NEFTune \cite{jain2023neftune}, which adds noise to create an augmented representation $\vec{x'} = \vec{x} + (\alpha/\sqrt{\lVert \vec{x} \rVert})\epsilon$, where $\alpha$ is the noise amplitude and $\epsilon = \text{Uniform}(-1,1)$. Following \citet{jain2023neftune}, we experiment with noise amplitudes from 0, 5, 10, 15, 20, and 25.

\vspace{-4pt}
\paragraph{Training Details} To address class imbalance, we perform a stratified dev-test split, creating a test set of 1,203 samples. Training employs stratified 10-fold cross-validation, maintaining class distributions across folds. We use oversampling on the training folds to balance classes. Models are trained for 60 epochs, with the best validation loss determining the final model. In total, we train 48 models,  covering all possible combinations of the four investigated variables (text encoder, use of text from profiles, use of text from retweets, and noise amplitude) to comprehensively assess their impact.

\vspace{-4pt}
\paragraph{Evaluation Metrics} We report F1 Macro, AUC Precision-Recall (AUC PR), and ROC AUC, in order of importance, chosen to address the dataset imbalance. F1 Macro offers a balanced evaluation by accounting for precision and recall across both classes, ensuring fair representation of major and minor classes. AUC PR is particularly suited for imbalanced datasets, emphasizing performance in detecting the positive class (fake news). ROC AUC evaluates the model's ability to distinguish between positive and negative classes, illustrating trade-offs between true and false positives. To rank the models, we applied the Wilcoxon-Holm post-hoc analysis of signed-rank paired differences, following methodology from \citet{demvsar2006statistical} and \citet{ferraz2021debacer}, ensuring statistical significance by comparing models trained on the same folds.

\section{Results and Discussion}

\begin{figure}
    \centering
    \includegraphics[width=\textwidth]{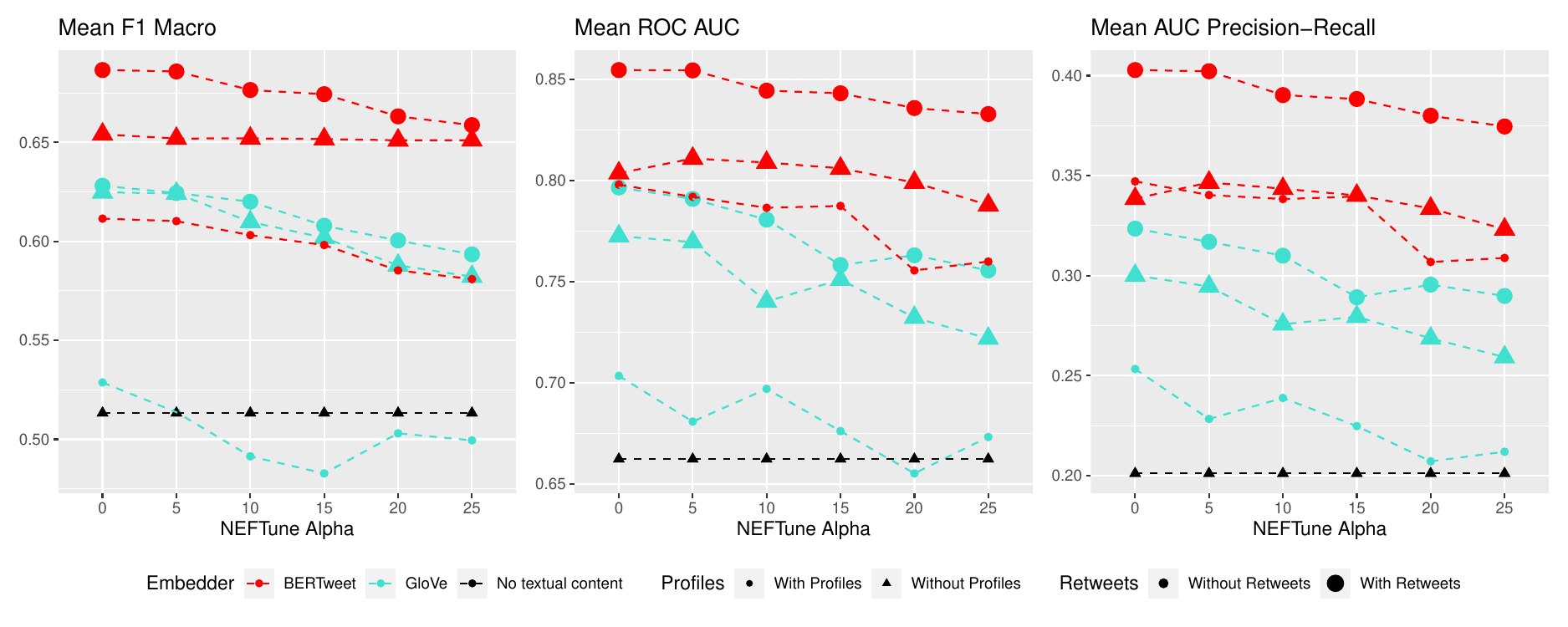}
    \caption{F1 Macro, ROC AUC and AUC PR as functions of Retweets, Profiles, Embedder, and NEFTune Alpha.}
    \label{fig:graficos_metricas}
    \vspace{-8pt}
\end{figure}

Figure~\ref{fig:graficos_metricas} presents the performance of different model configurations tested.
\vspace{-4pt}
\paragraph{Contextual Representations Significantly Enhance Performance.} For models without noise injection ($\alpha$ = 0), the contextual encoder BERTweet consistently outperformed static GloVe embeddings, confirming that contextual embeddings provide superior performance. Wilcoxon-Holm post-hoc analysis (p-values < 0.001) validated these results. The best configuration, incorporating profiles and retweets, improved Macro F1 by 9.3\% compared to GloVe and 33.8\% over models without textual features. These findings align with BERTweet's strengths in capturing context-specific nuances through its Transformer architecture, which, combined with its Twitter-specific pre-training, enables effective handling of informal social media language. This addresses RQ3, confirming the advantage of contextual embeddings over static ones and no text.
\vspace{-4pt}
\paragraph{Retweets Offer More Value than Profiles.} While incorporating textual content from both profiles and retweets improved model performance, retweet content provided significantly greater gains. This disparity is likely due to the sparse nature of user bios, which often contain limited or generic information, primarily aiding in identifying biased users or bots through keywords or the absence of content. In contrast, retweets offer richer, contextually relevant information directly tied to the propagated news, making them more effective for capturing meaningful patterns. Contextual text representations, such as those from BERTweet, are particularly effective at interpreting the complex relationships within retweet content—capabilities static embeddings often lack. Models using only profile data with GloVe performed similarly to those without textual features, and adding profiles alongside retweets did not improve GloVe configurations beyond those using retweets alone. This underscores GloVe's limitations in encoding relevant signals from profiles. Conversely, BERTweet successfully extracted useful information from both profiles and retweets, with the combination yielding additional performance gains. These findings address RQ1 and RQ2, demonstrating that while both profile and retweet textual features enhance performance, but retweet content provides substantially greater value.
\vspace{-4pt}
\paragraph{Noise Injection Compromises Model Performance and Stability.} Injecting noise into textual features consistently harmed performance across all metrics. Wilcoxon-Holm post-hoc analysis confirmed that models without noise outperformed those with noise in Macro F1, ROC AUC, and AUC PR, with 95\% confidence  (see Appendix \ref{appendix:other}). Increased noise amplitude caused instability during training, evidenced by higher variance in predictions. Rather than improving robustness, noise injection disrupted learning, hindering the model's ability to generalize effectively. Future work should explore methods to better balance noise injection and model learning for more effective data augmentation. This answers RQ4, showing that noise-injected oversampling does not improve performance and may hinder model stability.

\section{Conclusion and Future Work}

This work explored integrating textual information into GNNs for detecting disinformation campaigns. Incorporating contextual text representations into node features significantly enhanced GNN performance, yielding a 33.8\% relative gain in Macro F1 from model without text features. Combining retweet and profile content provided the best results, with retweets contributing more significantly to the gains. However, noise injection, despite its success in the text domain, proved unsuitable for GNNs, leading to instability and degraded performance. Future research should explore alternative data augmentation methods, such as structural manipulation, rather than relying solely on feature noise injection and oversampling.

\section*{Acknowledgements}

Bruno Croso Cunha da Silva and this research is funded by the \textit{Programa Unificado de Bolsas} (PUB) undergraduate research scholarship from Universidade de São Paulo, under project 3871/2023 "Detection of Fake News and Disinformation Campaigns via Natural Language Processing and Graph Analysis." We thank Nikos Kanakaris for his assistance in setting up the dataset for the initial experiments. We also thank Lucas Ribeiro da Silva, Sergio Magalhães Contente, Guilherme Mariano Francisco, and João Apolonio Matos, for their valuable aid with the Twitter API, the FakeNewsNet framework, and their collaboration during the early stages of this project. We extend our gratitude to the anonymous reviewers for their insightful feedback, which significantly contributed to improving this paper.

\bibliographystyle{unsrtnat}
\bibliography{reference}

\begin{thebibliography}{26}
\providecommand{\natexlab}[1]{#1}
\providecommand{\url}[1]{\texttt{#1}}
\expandafter\ifx\csname urlstyle\endcsname\relax
  \providecommand{\doi}[1]{doi: #1}\else
  \providecommand{\doi}{doi: \begingroup \urlstyle{rm}\Url}\fi

\bibitem[Shu et~al.(2019)Shu, Cui, Wang, Lee, and Liu]{dEFEND}
Kai Shu, Limeng Cui, Suhang Wang, Dongwon Lee, and Huan Liu.
\newblock defend: Explainable fake news detection.
\newblock In \emph{Proceedings of the 25th ACM SIGKDD international conference on knowledge discovery \& data mining}, pages 395--405, 2019.

\bibitem[Nasir et~al.(2021)Nasir, Khan, and Varlamis]{nasir2021fake}
Jamal~Abdul Nasir, Osama~Subhani Khan, and Iraklis Varlamis.
\newblock Fake news detection: A hybrid cnn-rnn based deep learning approach.
\newblock \emph{International Journal of Information Management Data Insights}, 1\penalty0 (1):\penalty0 100007, 2021.

\bibitem[Shu et~al.(2020{\natexlab{a}})Shu, Mahudeswaran, Wang, and Liu]{shu2020hierarchical}
Kai Shu, Deepak Mahudeswaran, Suhang Wang, and Huan Liu.
\newblock Hierarchical propagation networks for fake news detection: Investigation and exploitation.
\newblock In \emph{Proceedings of the international AAAI conference on web and social media}, volume~14, pages 626--637, 2020{\natexlab{a}}.

\bibitem[Scarselli et~al.(2009)Scarselli, Gori, Tsoi, Hagenbuchner, and Monfardini]{GNN_original2008}
Franco Scarselli, Marco Gori, Ah~Chung Tsoi, Markus Hagenbuchner, and Gabriele Monfardini.
\newblock The graph neural network model.
\newblock \emph{IEEE Transactions on Neural Networks}, 20\penalty0 (1):\penalty0 61--80, 2009.
\newblock \doi{10.1109/TNN.2008.2005605}.
\newblock URL \url{https://ieeexplore.ieee.org/abstract/document/4700287}.

\bibitem[Lu and Li(2020)]{gcan}
Yi-Ju Lu and Cheng-Te Li.
\newblock Gcan: Graph-aware co-attention networks for explainable fake news detection on social media.
\newblock In \emph{Proceedings of the 58th Annual Meeting of the Association for Computational Linguistics}, pages 505--514, 2020.

\bibitem[Hamid et~al.(2020)Hamid, Sheikh, Said, Ahmad, Gul, Hasan, and Al-Fuqaha]{hamid2020fake}
Abdullah Hamid, Nasrullah Sheikh, Naina Said, Kashif Ahmad, Asma Gul, Laiq Hasan, and Ala Al-Fuqaha.
\newblock Fake news detection in social media using graph neural networks and nlp techniques: A covid-19 use-case.
\newblock In \emph{Multimedia Evaluation Benchmark Workshop}. CEUR-WS, 2020.

\bibitem[Michail et~al.(2022)Michail, Kanakaris, and Varlamis]{michail2022detection}
Dimitrios Michail, Nikos Kanakaris, and Iraklis Varlamis.
\newblock Detection of fake news campaigns using graph convolutional networks.
\newblock \emph{International Journal of Information Management Data Insights}, 2\penalty0 (2):\penalty0 100104, 2022.

\bibitem[Devlin et~al.(2019)Devlin, Chang, Lee, and Toutanova]{bert2019}
Jacob Devlin, Ming-Wei Chang, Kenton Lee, and Kristina Toutanova.
\newblock {BERT}: Pre-training of deep bidirectional transformers for language understanding.
\newblock In Jill Burstein, Christy Doran, and Thamar Solorio, editors, \emph{Proceedings of the 2019 Conference of the North {A}merican Chapter of the Association for Computational Linguistics: Human Language Technologies, Volume 1 (Long and Short Papers)}, pages 4171--4186, Minneapolis, Minnesota, June 2019. Association for Computational Linguistics.
\newblock \doi{10.18653/v1/N19-1423}.
\newblock URL \url{https://aclanthology.org/N19-1423}.

\bibitem[Nguyen et~al.(2020)Nguyen, Vu, and Tuan~Nguyen]{nguyen2020bertweet}
Dat~Quoc Nguyen, Thanh Vu, and Anh Tuan~Nguyen.
\newblock {BERT}weet: A pre-trained language model for {E}nglish tweets.
\newblock In Qun Liu and David Schlangen, editors, \emph{Proceedings of the 2020 Conference on Empirical Methods in Natural Language Processing: System Demonstrations}, pages 9--14, Online, October 2020. Association for Computational Linguistics.
\newblock \doi{10.18653/v1/2020.emnlp-demos.2}.
\newblock URL \url{https://aclanthology.org/2020.emnlp-demos.2}.

\bibitem[Reimers and Gurevych(2019)]{reimers-gurevych-2019-sentence}
Nils Reimers and Iryna Gurevych.
\newblock Sentence-{BERT}: Sentence embeddings using {S}iamese {BERT}-networks.
\newblock In Kentaro Inui, Jing Jiang, Vincent Ng, and Xiaojun Wan, editors, \emph{Proceedings of the 2019 Conference on Empirical Methods in Natural Language Processing and the 9th International Joint Conference on Natural Language Processing (EMNLP-IJCNLP)}, pages 3982--3992, Hong Kong, China, November 2019. Association for Computational Linguistics.
\newblock \doi{10.18653/v1/D19-1410}.
\newblock URL \url{https://aclanthology.org/D19-1410}.

\bibitem[Jain et~al.(2024)Jain, yeh Chiang, Wen, Kirchenbauer, Chu, Somepalli, Bartoldson, Kailkhura, Schwarzschild, Saha, Goldblum, Geiping, and Goldstein]{jain2023neftune}
Neel Jain, Ping yeh Chiang, Yuxin Wen, John Kirchenbauer, Hong-Min Chu, Gowthami Somepalli, Brian~R. Bartoldson, Bhavya Kailkhura, Avi Schwarzschild, Aniruddha Saha, Micah Goldblum, Jonas Geiping, and Tom Goldstein.
\newblock {NEFT}une: Noisy embeddings improve instruction finetuning.
\newblock In \emph{The Twelfth International Conference on Learning Representations}, 2024.
\newblock URL \url{https://openreview.net/forum?id=0bMmZ3fkCk}.

\bibitem[Veličković et~al.(2018)Veličković, Cucurull, Casanova, Romero, Liò, and Bengio]{velickovic2018graph}
Petar Veličković, Guillem Cucurull, Arantxa Casanova, Adriana Romero, Pietro Liò, and Yoshua Bengio.
\newblock {Graph Attention Networks}.
\newblock In \emph{International Conference on Learning Representations}, 2018.
\newblock URL \url{https://openreview.net/forum?id=rJXMpikCZ}.

\bibitem[Ferraz et~al.(2024)Ferraz, Duarte, Ribeiro, Takayanagi, Alcoforado, Lopes, and Susi]{ferraz2024explainable}
Thomas~Palmeira Ferraz, Caio Henrique~Dias Duarte, Maria~Fernanda Ribeiro, Gabriel Goes~Braga Takayanagi, Alexandre Alcoforado, Roseli de~Deus Lopes, and Mart Susi.
\newblock {Explainable AI to Mitigate the Lack of Transparency and Legitimacy in Internet Moderation}.
\newblock \emph{Estudos Avan{\c{c}}ados}, 38:\penalty0 381--405, 2024.
\newblock URL \url{https://www.scielo.br/j/ea/a/KPMcWYkkqHy5ZK3zTFCBpFj/?lang=en}.

\bibitem[Liu et~al.(2022)Liu, Dong, Li, Xu, Rong, Zhao, Huang, and Wu]{liu2022local}
Songtao Liu, Hanze Dong, Lanqing Li, Tingyang Xu, Yu~Rong, Peilin Zhao, Junzhou Huang, and Dinghao Wu.
\newblock Local augmentation for graph neural networks, 2022.
\newblock URL \url{https://openreview.net/forum?id=3FvF1db-bKT}.

\bibitem[Gao and Ji(2019)]{gao2019graph}
Hongyang Gao and Shuiwang Ji.
\newblock Graph u-nets.
\newblock In \emph{international conference on machine learning}, pages 2083--2092. PMLR, 2019.

\bibitem[Zhou and Gong(2023)]{zhou2023graphsr}
Mengting Zhou and Zhiguo Gong.
\newblock Graphsr: a data augmentation algorithm for imbalanced node classification.
\newblock In \emph{Proceedings of the AAAI Conference on Artificial Intelligence}, volume~37, pages 4954--4962, 2023.

\bibitem[Zhao et~al.(2024)Zhao, Zhang, and Wang]{zhao2024imbalanced}
Tianxiang Zhao, Xiang Zhang, and Suhang Wang.
\newblock Imbalanced node classification with synthetic over-sampling.
\newblock \emph{IEEE Transactions on Knowledge and Data Engineering}, 2024.

\bibitem[You et~al.(2020)You, Chen, Sui, Chen, Wang, and Shen]{you2020graph}
Yuning You, Tianlong Chen, Yongduo Sui, Ting Chen, Zhangyang Wang, and Yang Shen.
\newblock Graph contrastive learning with augmentations.
\newblock \emph{Advances in neural information processing systems}, 33:\penalty0 5812--5823, 2020.

\bibitem[Kong et~al.(2021)Kong, Li, Ding, Wu, Zhu, Ghanem, Taylor, and Goldstein]{kong2021flag}
Kezhi Kong, Guohao Li, Mucong Ding, Zuxuan Wu, Chen Zhu, Bernard Ghanem, Gavin Taylor, and Tom Goldstein.
\newblock {\{}FLAG{\}}: Adversarial data augmentation for graph neural networks, 2021.
\newblock URL \url{https://openreview.net/forum?id=mj7WsaHYxj}.

\bibitem[Liang and Liang(2024)]{liang2024drbert}
Wen Liang and Youzhi Liang.
\newblock Drbert: Unveiling the potential of masked language modeling decoder in bert pretraining.
\newblock \emph{arXiv preprint arXiv:2401.15861}, 2024.

\bibitem[Chen et~al.(2024)Chen, Han, Misra, Li, Hu, Zou, Sugiyama, Wang, and Raj]{chen2024slight}
Hao Chen, Yujin Han, Diganta Misra, Xiang Li, Kai Hu, Difan Zou, Masashi Sugiyama, Jindong Wang, and Bhiksha Raj.
\newblock Slight corruption in pre-training data makes better diffusion models.
\newblock \emph{arXiv preprint arXiv:2405.20494}, 2024.

\bibitem[Yang et~al.(2024)Yang, Huang, Tan, and Wang]{classificationneftune2024}
Liziqiu Yang, Yanhao Huang, Cong Tan, and Sen Wang.
\newblock News topic classification base on fine-tuning of chatglm3-6b using neftune and lora.
\newblock In \emph{Proceedings of the 2024 International Conference on Computer and Multimedia Technology}, ICCMT '24, page 521–525, New York, NY, USA, 2024. Association for Computing Machinery.
\newblock ISBN 9798400718267.
\newblock \doi{10.1145/3675249.3675339}.
\newblock URL \url{https://doi.org/10.1145/3675249.3675339}.

\bibitem[Shu et~al.(2020{\natexlab{b}})Shu, Mahudeswaran, Wang, Lee, and Liu]{shu2020fakenewsnet}
Kai Shu, Deepak Mahudeswaran, Suhang Wang, Dongwon Lee, and Huan Liu.
\newblock Fakenewsnet: A data repository with news content, social context, and spatiotemporal information for studying fake news on social media.
\newblock \emph{Big data}, 8\penalty0 (3):\penalty0 171--188, 2020{\natexlab{b}}.

\bibitem[Pennington et~al.(2014)Pennington, Socher, and Manning]{pennington2014glove}
Jeffrey Pennington, Richard Socher, and Christopher~D Manning.
\newblock Glove: Global vectors for word representation.
\newblock In \emph{Proceedings of the 2014 conference on empirical methods in natural language processing (EMNLP)}, pages 1532--1543, 2014.

\bibitem[Dem{\v{s}}ar(2006)]{demvsar2006statistical}
Janez Dem{\v{s}}ar.
\newblock Statistical comparisons of classifiers over multiple data sets.
\newblock \emph{Journal of Machine Learning Research}, 7:\penalty0 1--30, 2006.

\bibitem[Ferraz et~al.(2021)Ferraz, Alcoforado, Bustos, Oliveira, Gerber, M{\"u}ller, d’Almeida, Veloso, and Costa]{ferraz2021debacer}
Thomas~Palmeira Ferraz, Alexandre Alcoforado, Enzo Bustos, Andr{\'e}~Seidel Oliveira, Rodrigo Gerber, Na{\'\i}de M{\"u}ller, Andr{\'e}~Corr{\^e}a d’Almeida, Bruno~Miguel Veloso, and Anna Helena~Reali Costa.
\newblock Debacer: a method for slicing moderated debates.
\newblock In \emph{ENIAC 2021: XVIII encontro nacional de intelig{\^e}ncia artificial e computacional. 18th national meeting on artificial and computational intelligence}, pages 667--678. Sociedade Brasileira de Computa{\c{c}}{\~a}o, 2021.

\end{thebibliography}

\newpage
\appendix
\section{Other experimental results}
\label{appendix:other}

\begin{figure}[h]
    \centering
    \includegraphics[width=\textwidth]{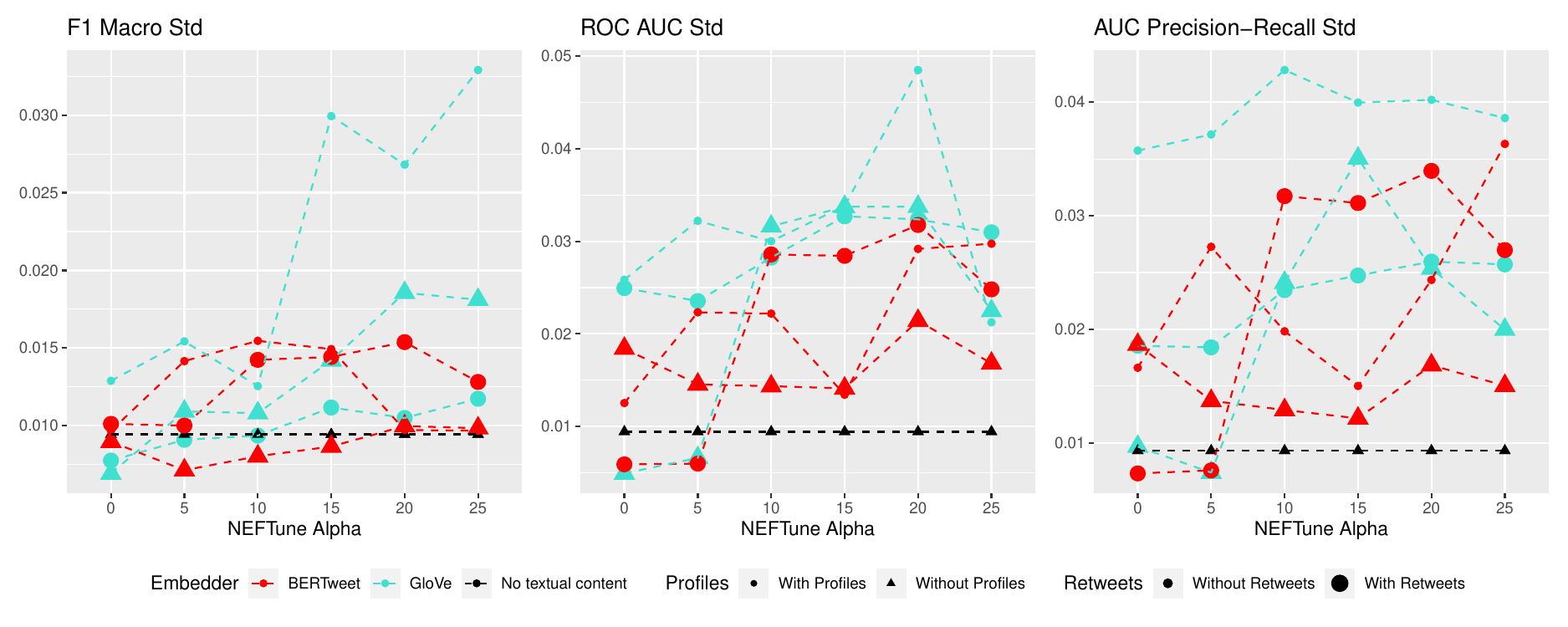}
        \caption{F1 Macro, ROC AUC and AUC PR standard deviations as functions of Retweets, Profiles, Embedder, and NEFTune Alpha.}
    \label{fig:graficos_metricas_std}
\end{figure}

Figure~\ref{fig:graficos_metricas_std} present the standard deviation on the models studied. An increasing trend in the standard deviation was observed as more noise was added, indicating that, as expected, the model’s predictions become more unstable with the increase of noise. For the other features, no clear patterns were observed. However, it is important to emphasize the need for future studies focused on verifying the statistical significance of the aforementioned conclusions.

The mean and standard deviations of F1 Macro, ROC AUC and AUC PR of each trained model, used to construct Figures~\ref{fig:graficos_metricas} and \ref{fig:graficos_metricas_std}, are presented in Table~\ref{tab:complete_results}.

\begin{table}[h]
\centering
\caption{P-values obtained from the pairwise comparison of NEFTune alphas for each metric}
\label{tab:p_values_neftune}
\begin{tabular}{cccc}
\toprule
\multirow{2.2}{*}{NEFTune Alpha} & \multicolumn{3}{c}{Metric} \\
\cmidrule(lr){2-4}
 & F1 Macro & ROC AUC & AUC PR \\
 \midrule 
0 vs 5 & 0.051& 0.048& 0.059\\
0 vs 10 & <0.001 & 0.010& 0.011\\
0 vs 15 & <0.001 & <0.001 & 0.001\\
0 vs 20 & <0.001 & <0.001 & <0.001 \\
0 vs 25 & <0.001 & <0.001 & <0.001 \\
5 vs 10 & <0.001 & 0.058& 0.059\\
5 vs 15 & <0.001 & <0.001& 0.007\\
5 vs 20 & <0.001 & <0.001 & <0.001 \\
5 vs 25 & <0.001 & <0.001 & <0.001 \\
10 vs 15 & 0.008& 0.025& 0.059\\
10 vs 20 & 0.003& <0.001 & <0.001 \\
10 vs 25 & <0.001 & <0.001 & <0.001 \\
15 vs 20 & 0.051& 0.003& 0.004\\
15 vs 25 & 0.004& <0.001 & 0.001\\
20 vs 25 & 0.051& 0.084& 0.059\\
\bottomrule

\end{tabular}

\end{table}

\begin{table}
\centering
\caption{Mean and standard deviation, in percentage, of F1 Macro, ROC AUC and AUC Precision-Recall for each trained model}
\label{tab:complete_results}
\begin{tabular}{c c c c c c c}
\toprule
\textbf{Encoder} & \textbf{Profiles} & \textbf{Retweets} & \textbf{NEFTune Alpha} & \textbf{F1 Macro} & \textbf{ROC AUC} & \textbf{AUC PR} \\
\midrule
GloVe & Absent & Absent & 0 & 51.3 ± 0.9 & 66.2 ± 0.9 & 20.1 ± 0.9 \\
GloVe & Absent & Absent & 5 & 51.3 ± 0.9 & 66.2 ± 0.9 & 20.1 ± 0.9 \\
GloVe & Absent & Absent & 10 & 51.3 ± 0.9 & 66.2 ± 0.9 & 20.1 ± 0.9 \\
GloVe & Absent & Absent & 15 & 51.3 ± 0.9 & 66.2 ± 0.9 & 20.1 ± 0.9 \\
GloVe & Absent & Absent & 20 & 51.3 ± 0.9 & 66.2 ± 0.9 & 20.1 ± 0.9 \\
GloVe & Absent & Absent & 25 & 51.3 ± 0.9 & 66.2 ± 0.9 & 20.1 ± 0.9 \\
GloVe & Present & Absent & 0 & 52.9 ± 1.3 & 70.3 ± 2.6 & 25.3 ± 3.6 \\
GloVe & Present & Absent & 5 & 51.4 ± 1.5 & 68.1 ± 3.2 & 22.8 ± 3.7 \\
GloVe & Present & Absent & 10 & 49.1 ± 1.3 & 69.7 ± 3.0 & 23.9 ± 4.3 \\
GloVe & Present & Absent & 15 & 48.3 ± 3.0 & 67.6 ± 3.4 & 22.5 ± 4.0 \\
GloVe & Present & Absent & 20 & 50.3 ± 2.7 & 65.5 ± 4.9 & 20.7 ± 4.0 \\
GloVe & Present & Absent & 25 & 49.9 ± 3.3 & 67.3 ± 2.1 & 21.2 ± 3.9 \\
GloVe & Absent & Present & 0 & 62.5 ± 0.7 & 77.2 ± 0.5 & 30.0 ± 1.0 \\
GloVe & Absent & Present & 5 & 62.4 ± 1.1 & 77.0 ± 0.7 & 29.5 ± 0.7 \\
GloVe & Absent & Present & 10 & 61.0 ± 1.1 & 74.0 ± 3.2 & 27.6 ± 2.4 \\
GloVe & Absent & Present & 15 & 60.2 ± 1.4 & 75.1 ± 3.4 & 27.9 ± 3.5 \\
GloVe & Absent & Present & 20 & 58.8 ± 1.9 & 73.2 ± 3.4 & 26.9 ± 2.5 \\
GloVe & Absent & Present & 25 & 58.2 ± 1.8 & 72.2 ± 2.2 & 25.9 ± 2.0 \\
GloVe & Present & Present & 0 & 62.8 ± 0.8 & 79.7 ± 2.5 & 32.4 ± 1.9 \\
GloVe & Present & Present & 5 & 62.4 ± 0.9 & 79.1 ± 2.4 & 31.7 ± 1.8 \\
GloVe & Present & Present & 10 & 62.0 ± 0.9 & 78.1 ± 2.8 & 31.0 ± 2.3 \\
GloVe & Present & Present & 15 & 60.8 ± 1.1 & 75.8 ± 3.3 & 28.9 ± 2.5 \\
GloVe & Present & Present & 20 & 60.0 ± 1.0 & 76.3 ± 3.2 & 29.5 ± 2.6 \\
GloVe & Present & Present & 25 & 59.3 ± 1.2 & 75.6 ± 3.1 & 29.0 ± 2.6 \\
BERTweet & Absent & Absent & 0 & 51.3 ± 0.9 & 66.2 ± 0.9 & 20.1 ± 0.9 \\
BERTweet & Absent & Absent & 5 & 51.3 ± 0.9 & 66.2 ± 0.9 & 20.1 ± 0.9 \\
BERTweet & Absent & Absent & 10 & 51.3 ± 0.9 & 66.2 ± 0.9 & 20.1 ± 0.9 \\
BERTweet & Absent & Absent & 15 & 51.3 ± 0.9 & 66.2 ± 0.9 & 20.1 ± 0.9 \\
BERTweet & Absent & Absent & 20 & 51.3 ± 0.9 & 66.2 ± 0.9 & 20.1 ± 0.9 \\
BERTweet & Absent & Absent & 25 & 51.3 ± 0.9 & 66.2 ± 0.9 & 20.1 ± 0.9 \\
BERTweet & Present & Absent & 0 & 61.1 ± 1.0 & 79.8 ± 1.3 & 34.7 ± 1.7 \\
BERTweet & Present & Absent & 5 & 61.0 ± 1.4 & 79.2 ± 2.2 & 34.0 ± 2.7 \\
BERTweet & Present & Absent & 10 & 60.3 ± 1.5 & 78.6 ± 2.2 & 33.8 ± 2.0 \\
BERTweet & Present & Absent & 15 & 59.8 ± 1.5 & 78.7 ± 1.3 & 34.0 ± 1.5 \\
BERTweet & Present & Absent & 20 & 58.5 ± 1.0 & 75.6 ± 2.9 & 30.7 ± 2.4 \\
BERTweet & Present & Absent & 25 & 58.1 ± 1.0 & 76.0 ± 3.0 & 30.9 ± 3.6 \\
BERTweet & Absent & Present & 0 & 65.4 ± 0.9 & 80.4 ± 1.8 & 33.9 ± 1.9 \\
BERTweet & Absent & Present & 5 & 65.2 ± 0.7 & 81.1 ± 1.5 & 34.7 ± 1.4 \\
BERTweet & Absent & Present & 10 & 65.2 ± 0.8 & 80.9 ± 1.4 & 34.3 ± 1.3 \\
BERTweet & Absent & Present & 15 & 65.2 ± 0.9 & 80.6 ± 1.4 & 34.0 ± 1.2 \\
BERTweet & Absent & Present & 20 & 65.1 ± 1.0 & 79.9 ± 2.1 & 33.4 ± 1.7 \\
BERTweet & Absent & Present & 25 & 65.1 ± 1.0 & 78.8 ± 1.7 & 32.3 ± 1.5 \\
BERTweet & Present & Present & 0 & 68.7 ± 1.0 & 85.5 ± 0.6 & 40.3 ± 0.7 \\
BERTweet & Present & Present & 5 & 68.6 ± 1.0 & 85.4 ± 0.6 & 40.2 ± 0.8 \\
BERTweet & Present & Present & 10 & 67.6 ± 1.4 & 84.4 ± 2.9 & 39.0 ± 3.2 \\
BERTweet & Present & Present & 15 & 67.4 ± 1.4 & 84.3 ± 2.8 & 38.8 ± 3.1 \\
BERTweet & Present & Present & 20 & 66.3 ± 1.5 & 83.6 ± 3.2 & 38.0 ± 3.4 \\
BERTweet & Present & Present & 25 & 65.9 ± 1.3 & 83.3 ± 2.5 & 37.5 ± 2.7 \\
\bottomrule
\end{tabular}

\end{table}

The p-values obtained from the comparison of each pair of NEFTune amplitude for each metric are presented in Table~\ref{tab:p_values_neftune}. The observations made in the descriptive analysis were corroborated by paired Wilcoxon tests, which showed that models without noise have superior F1 Macro and ROC AUC metrics, with a 95\% confidence level, compared to models with noise. For the AUC PR metric, the null hypothesis of no difference between the group medians was not rejected when comparing models without noise to those with NEFTune noise alpha of 5, 10, and 15, but it was rejected for higher noise levels.

\end{document}